\documentclass[runningheads]{llncs}
\usepackage{graphicx}
\usepackage{amsmath,amssymb} 
\usepackage{color}
\usepackage[width=122mm,left=12mm,paperwidth=146mm,height=193mm,top=12mm,paperheight=217mm]{geometry}

\usepackage{cvpr}
\makeatletter
\@namedef{ver@everyshi.sty}{}
\makeatother
\usepackage{times}
\usepackage{epsfig}
\usepackage[pagebackref=true,breaklinks=true,letterpaper=true,colorlinks,bookmarks=false]{hyperref}
\usepackage{pgffor}
\usepackage{dsfont}
\usepackage{subcaption}

\DeclareMathOperator*{\argmin}{arg\,min}

\newcommand*{\affaddr}[1]{#1} 
\newcommand*{\affmark}[1][*]{\textsuperscript{#1}}

\begin{document}
\pagestyle{headings}
\mainmatter

\title{Semi-supervised Adversarial Learning to Generate Photorealistic Face Images of New Identities from 3D Morphable Model} 

\titlerunning{Semi-supervised Adv. Learning to Generate Face Images of New Ids from 3DMM}

\authorrunning{B. Gecer,  B. Bhattarai, J. Kittler, and  T.K. Kim} 

\author{
Baris Gecer\affmark[1], Binod Bhattarai\affmark[1], Josef Kittler\affmark[2], and Tae-Kyun Kim\affmark[1]
}

\institute{\affaddr{\affmark[1]Department of Electrical and Electronic Engineering,
Imperial College London, UK}\\
\affaddr{\affmark[2]Centre for Vision, Speech and Signal Processing, University of Surrey, UK}\\
\email{ \{b.gecer,b.bhattarai,tk.kim\}@imperial.ac.uk, j.kittler@surrey.ac.uk}}


\maketitle
\begin{abstract}
We propose a novel end-to-end semi-supervised adversarial framework to generate photorealistic face images of new identities with wide ranges of expressions, poses, and illuminations conditioned by a 3D morphable model. Previous adversarial style-transfer methods either supervise their networks with large volume of paired data or use unpaired data with a highly under-constrained two-way generative framework in an unsupervised fashion. We introduce pairwise adversarial supervision to constrain two-way domain adaptation by a small number of paired real and synthetic images for training along with the large volume of unpaired data. Extensive qualitative and quantitative experiments are performed to validate our idea. Generated face images of new identities contain pose, lighting and expression diversity and qualitative results show that they are highly constraint by the synthetic input image while adding photorealism and retaining identity information. We combine face images generated by the proposed method with the real data set to train face recognition algorithms. We evaluated the model on two challenging data sets: LFW and IJB-A. We observe that the generated images from our framework consistently improves over the performance of deep face recognition network trained with Oxford VGG Face dataset and achieves comparable results to the state-of-the-art.
 \end{abstract}

%

\section{Introduction}
\label{sec:intro}
Deep learning has shown an great improvement in performance of several computer vision tasks~\cite{ren2015faster,he2017mask,geccer2016detection,dong2016image,dosovitskiy2015flownet,yuan2017bighand2} including face recognition~\cite{parkhi2015deep,schroff2015facenet,xiong2015conditional,liu2017sphereface,xiong2016convolutional} in the recent years. This was mainly thanks to the availability of large-scale datasets. Yet the performance is often limited by the volume and the variations of training examples. Larger and wider datasets usually improve the generalization and overall performance of the model \cite{schroff2015facenet,bansal2017s}.

The process of collecting and annotating training examples for every specific computer vision task is laborious and non-trivial. To overcome this challenge, additional synthetic training examples along with limited real training examples can be utilised to train the model. Some of the recent works such as 3D face reconstruction~\cite{richardson20163d}, gaze estimation~\cite{zhang2015appearance,wood2016learning}, human pose, shape and motion estimation~\cite{varol2017learning} \etc use  additional synthetic images generated from 3D models to train deep networks. 
One can generate synthetic face images using a 3D morphable model (3DMM) \cite{blanz1999morphable} by manipulating identity, expression, illumination, and pose parameters. However, the resulting images are not photorealistic enough to be suitable for in-the-wild face recognition tasks. It is beacause the information of real face scans is compressed by the 3DMM and the graphical engine that models illumination and surface is not perfectly accurate. Thus, the main challenge of using synthetic data obtained from 3DMM model is the discrepancy in nature and quality of synthetic and real images which pose the problem of domain adaptation~\cite{patel2015visual}. Recently, adversarial training methods~\cite{shrivastava2016learning,sixt2016rendergan,costa2017towards} become popular to mitigate such challenges.



\begin{figure}[t]
\includegraphics[width=\textwidth]{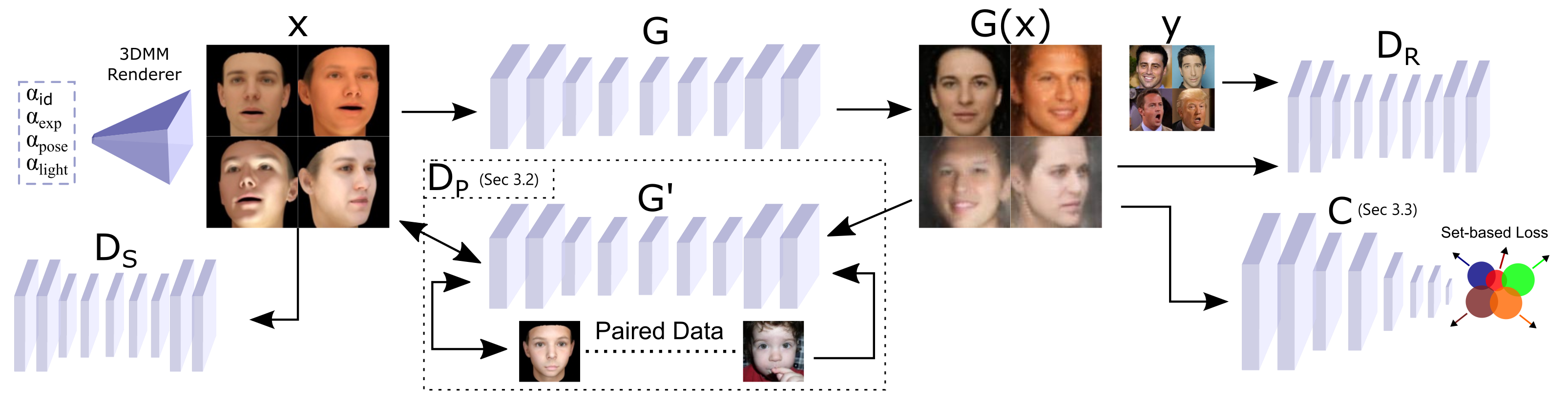}
\caption{Our approach aims to synthesize photorealistic images conditioned by a given synthetic image by 3DMM. It regularizes cycle consistency \cite{zhu2017unpaired} by introducing an additional adversarial game between the two generator networks in an unsupervised fashion. Thus the under-constraint cycle loss is supervised to have correct matching between the two domains by the help of a limited number of paired data. We also encourage the generator to preserve face identity by a set-based supervision through a pretrained classification network.}
\label{fig:overview}
\end{figure}

Generative Adversarial Network (GAN), introduced by Goodfellow \etal~\cite{goodfellow2016nips}, and its variants~\cite{radford2015unsupervised,karras2017progressive,berthelot2017began,dumoulin2016adversarially} 
are quite successful in generating realistic images. However, in practice, GANs are likely to stuck in mode collapse for large scale image generation. They are also unable to produce images that are 3D coherent and globally consistent~\cite{goodfellow2016nips}. To overcome these drawbacks, we propose a semi-supervised adversarial learning 
framework to synthesize photorealistic face images of new identities with numerous data variation supplied by a 3DMM. We address these shortcomings by exciting a generator 
network with synthetic images sampled from 3DMM and transforming them into photorealistic domain using adversarial training as a bridge. Unlike most of the existing works 
that excite their generators with a noise vector \cite{radford2015unsupervised,berthelot2017began}, we feed our generator network by synthetic face images. Such a strong constraint naturally helps in avoiding the mode collapse problem, one of the main challenges faced by the current GAN methods. Fig.~\ref{fig:overview} shows the general overview of the proposed method. We discuss the proposed method in more details in Sec.~\ref{sec:approach}.



In this paper, we address the challenge of generating photorealistic 
face images from 3DMM rendered faces of different identities with arbitrary poses, expressions, and illuminations. We formulate this problem as a domain adaptation 
problem \ie aligning  the 3DMM rendered face domain into realistic face domain. One of the previous works closest to ours ~\cite{isola2016image} address style transfer 
problem between a pair of domains with classical conditional GAN. The major bottleneck of this method is, it requires a large number of  paired examples from both 
domains which are hard to collect. CycleGAN \cite{zhu2017unpaired}, another recent method and closest to our work, proposes a two-way GAN framework for unsupervised image-to-image translation. However, the cycle consistency loss proposed in their method is satisfied as long as the transitivity of the two mapping networks is maintained.
Thus, the resulting mapping is not guaranteed to produce the intended transformation. To overcome the drawbacks of these methods~\cite{isola2016image,zhu2017unpaired}, 
we propose to use a small amount of paired data to train an inverse mapping network as a matching aware discriminator. In the proposed method, the inverse mapping network  
plays the role of both the generator and the discriminator. To the best of our knowledge, this is the first attempt for adversarial semi-supervised style translation for an 
application with such limited paired data.



Adding realism to the synthetic face images and preserving their identity information is a challenging problem. 
Although synthetic input images, 3DMM rendered faces, contain distinct face identities, the distinction between them vanishes as a result of the virtue of non-linear transformations while the discriminator encourages realism. 
To tackle such problem, prior works either employ a separate pre-trained network~\cite{yin2017towards} or embed Identity labels (id)~\cite{tran2017disentangled} into 
the discriminator. Unlike existing works, which are focused on generating new images of existing identities, we are interested in generating multiple images of
new identities itself. Therefore, such techniques are not directly applicable to our problem. To address this challenge, we propose to use set-based center~\cite{wen2016discriminative} and pushing loss 
functions~\cite{gecer2017learning} on top of a pre-trained face embedding network. This will keep track of the changing average of embeddings of generated 
images belonging to same identity (i.e. centroids). In this way identity preservation becomes adaptive to changing feature space during the training of the generator network unlike softmax layer that converges very quickly at the beginning of the training before meaningful images are generated.




Our contributions can be summarized as follows:
\begin{itemize}
\item We propose a novel end-to-end adversarial training framework to generate photorealistic face images of new identities constrained by synthetic 3DMM images with identity, pose, illumination and expression diversity. The resulting synthetic face images are visually plausible and can be used to boost face recognition as additional training data or any other graphical purposes.
\item We propose a novel semi-supervised adversarial style transfer approach that trains an inverse mapping network as a discriminator with paired synthetic-real images.
\item We employ a novel set-based loss function to preserve consistency among unknown identities during GAN training.
\end{itemize}

\section{Related Works}
\label{sec:rltd_works}
In this Section we discuss the prior art that is closely related to the proposed method.

\paragraph{Domain Adaptation.} As stated in the introduction, our problem of generating photorealistic face images from 3DMM rendered faces can be seen as a domain adaptation problem. A straightforward adaptation approach is to align the distributions at the feature level by simply adding a loss to measure the mismatch either through second-order moments~\cite{sun2015subspace} or with adversarial losses~\cite{tzeng2015simultaneous,tzeng2017adversarial,ganin2016domain}. 

Recently, pixel level domain adaptation becomes popular due to practical breakthroughs on Kullback-Leibler divergence~\cite{goodfellow2014generative,goodfellow2016nips,radford2015unsupervised}, namely GANs which optimize a generative and discriminative network through a mini-max game. It has been applied to a wide range of applications including fashion clothing~\cite{lassner2017generative}, person specific avatar creation~\cite{wolf2017unsupervised}, text-to-image synthesis~\cite{zhang2016stackgan}, face frontalization~\cite{yin2017towards}, and retinal image synthesis~\cite{costa2017towards}. 

Pixel domain adaptation can be done in a supervised manner simply by conditioning the discriminator network~\cite{isola2016image} or directly the output of the generator~\cite{chen2017photographic} with the expected output when there is enough paired data from both domains. Please note collecting a large number of paired training examples is expensive, and often requires expert knowledge.~\cite{reed2016generative} proposes a text-to-image synthesis GAN with a matching aware discriminator. They optimize their discriminator for image-text matching beside requiring realism with an additional mismatched text-image pair.

For the cases where paired data is not available, many approaches take an unsupervised way such as pixel-level consistency between input and output of the generator network~\cite{bousmalis2016unsupervised,shrivastava2016learning}, an encoder architecture that is shared by both domains\cite{bousmalis2016domain} and adaptive instance normalization~\cite{huang2017arbitrary}. An interesting approach is to have two way translation between domains with two distinct generator and discriminator networks. They  constrain the two mappings to be inverses of each other with either ResNet~\cite{zhu2017unpaired} or encoder-decoder network~\cite{liu2017unsupervised} as the generator.

\paragraph{Synthetic Training Data Generation.}
The usage of synthetic data as additional training data is shown to be helpful even if they are graphically rendered images in many applications such as 3D face reconstruction~\cite{richardson20163d}, gaze estimation ~\cite{zhang2015appearance,wood2016learning}, human pose, shape and motion estimation~\cite{varol2017learning}. Despite the availability of almost infinite number of synthetic images, those approaches are limited due to the domain difference from that of in-the-wild images. 

Many existing works utilized adversarial domain adaptation to translate images into photorealistic domain such that they are more useful as a training data.~\cite{zheng2017unlabeled} generates many unlabeled samples to improve person re-identification in a semi-supervised fashion. RenderGAN~\cite{sixt2016rendergan} proposes a sophisticated approach to refine graphically rendered synthetic images of tagged bees to be used as training data for bee tag decoding application. WaterGAN~\cite{li2017watergan} synthesizes realistic underwater images by modeling camera parameters and environment effects explicitly to be used as training data for color correction task. Some studies deform existing images by a 3D model to augment diverse set of dataset~\cite{masi2016we} without adversarial learning.

One of the recent works, simGAN~\cite{shrivastava2016learning}, generates realistic synthetic data to improve eye gaze and hand pose estimation. It optimizes pixel level correspondence between input and output of the generator network to preserve content of the synthetic image. This is in fact a limited solution since the pixel-consistency loss encourages the generated images to be similar to synthetic input images and it partially contradicts adversarial realism loss. Instead, we employ an inverse translation network similar to cycleGAN~\cite{zhu2017unpaired} with an additional pair-wise supervision to preserve the initial condition without hurting realism. This network also behaves as a discriminator to a straight mapping network with a real paired data to avoid possible biased translation.

\paragraph{Identity Preservation.}
To preserve the identity/category of the synthesized images, some of the recent works such as~\cite{chen2016infogan,tran2017disentangled} keep categorical/identity information in discriminator network as an additional task. 
Some of the others propose to employ a separate classification network which is usually pre-trained~\cite{lu2017conditional,yin2017towards}. In both these cases, the categories/identities are known beforehand and are fixed in number. Thus it is trivial to  include such supervision in a GAN framework by training the classifier with real data. However such setup is not feasible in our case as images of new identities to-be-generated are not available to pre-train a classification network (see Section \ref{sec:id_preservation} for further discussion)

To address the limitation of existing methods of retaining identity/category information of synthesized images, we employ a combination of different set-based supervision approaches for unknown identities to be distinct in the pre-trained embedding space. We keep track of moving averages of same-id features by the momentum-like centroid update rule of center loss~\cite{wen2016discriminative} and penalize distant same-id samples and close different-id samples by a simplified variant of magnet loss\cite{rippel2015metric} without its sampling process and with only one cluster per identity.



\section{Adversarial Identity Generation}
\label{sec:approach}


In this Section, we describe in details the proposed method. Fig.~\ref{fig:overview} shows the detailed schematic diagram of our method. Specifically, the synthetic image set $x \in \mathcal{S}$ is formed by a graphical engine for the randomly sampled 3DMM, pose and lighting parameters $\alpha$. Then they are translated into more photorealistic domain $G(x)$ through the network $G$ and mapped back to synthetic domain ($G'(G(x))$) through the network $G'$ to retain $x$. Adversarial synthetic and real domain translation of $G$ and $G'$ networks are supervised by the discriminator networks $D_R$ and $D_S$, with an additional adversarial game between $G$ and $G'$ as generator and discriminator respectively. During training, generated identities by 3DMM is preserved with a set-based loss on a pre-trained embedding network $C$.
In the following sub-sections, we further describe these components \ie domain adaptation, real-synthetic pair discriminator, and identity preservation.

\subsection{Unsupervised Domain Adaptation}
Given a 3D morphable model (3DMM)~\cite{blanz1999morphable}, we synthesize face images of new identities sampled from its Principal Components Analysis (PCA) coefficients' space with random variation of expression, lighting and pose. Similar to~\cite{zhu2017unpaired}, a synthetic input image ($x \in \mathcal{S}$) is mapped to photorealistic domain by a residual network ($G : S \rightarrow \hat{R}$) and mapped back to synthetic domain by a 3DMM fitting network ($G' : \hat{R} \rightarrow \hat{S}$) to complete forward cycle only. To preserve cycle consistency, the resulting image $G'(G(x))$ is encouraged to be the same as input $x$ by a pixel level $L_1$ loss:
\begin{align}
\mathcal{L}_{cyc} &= \mathds{E}_{x \in \mathcal{S}}\|G'(G(x))-x\|_1
\end{align}
In order to encourage the resulting images $G(x)$ and $G'(G(x))$ to have similar distribution as real and synthetic domains respectively, those refiner networks are supervised by discriminator networks $D_R$ and $D_S$ with images of the respective domains. The discriminator networks are formed as auto-encoders as in boundary equilibrium GAN (BEGAN) architecture \cite{berthelot2017began} in which the generator and discriminator networks are trained by the following adversarial training formulation:

\begin{align}
\mathcal{L}_{G} &= \mathds{E}_{x \in \mathcal{S}} \|G(x)-D_R(G(x))\|_1 \\
\mathcal{L}_{G'} &= \mathds{E}_{x \in \mathcal{S}} \|G'(G(x))-D_S(G'(G(x)))\|_1 \\
\mathcal{L}_{D_R} &= \mathds{E}_{x \in \mathcal{S},y \in \mathcal{R}} \|y-D_R(y)\|_1 -k_t^{D_R} \mathcal{L}_{G} \\
\mathcal{L}_{D_S} &= \mathds{E}_{x \in \mathcal{S}} \|x-D_S(x)\|_1 -k_t^{D_S} \mathcal{L}_{G'} 
\end{align}
where for each training step $t$ and the network $G$ we update the balancing term with $k_t^{D,G} = k_{t-1}^{D,G} + 0.001(0.5\mathcal{L}_{D} - \mathcal{L}_{G})$. As suggested by~\cite{berthelot2017began}, this term helps to balance between generator and discriminator and stabilize the training.

\subsection{Adversarial Pair Matching}

\begin{figure*}[t] 
\centering   
\begin{subfigure}[t]{0.24\textwidth}
\centering
\includegraphics[width=\textwidth]{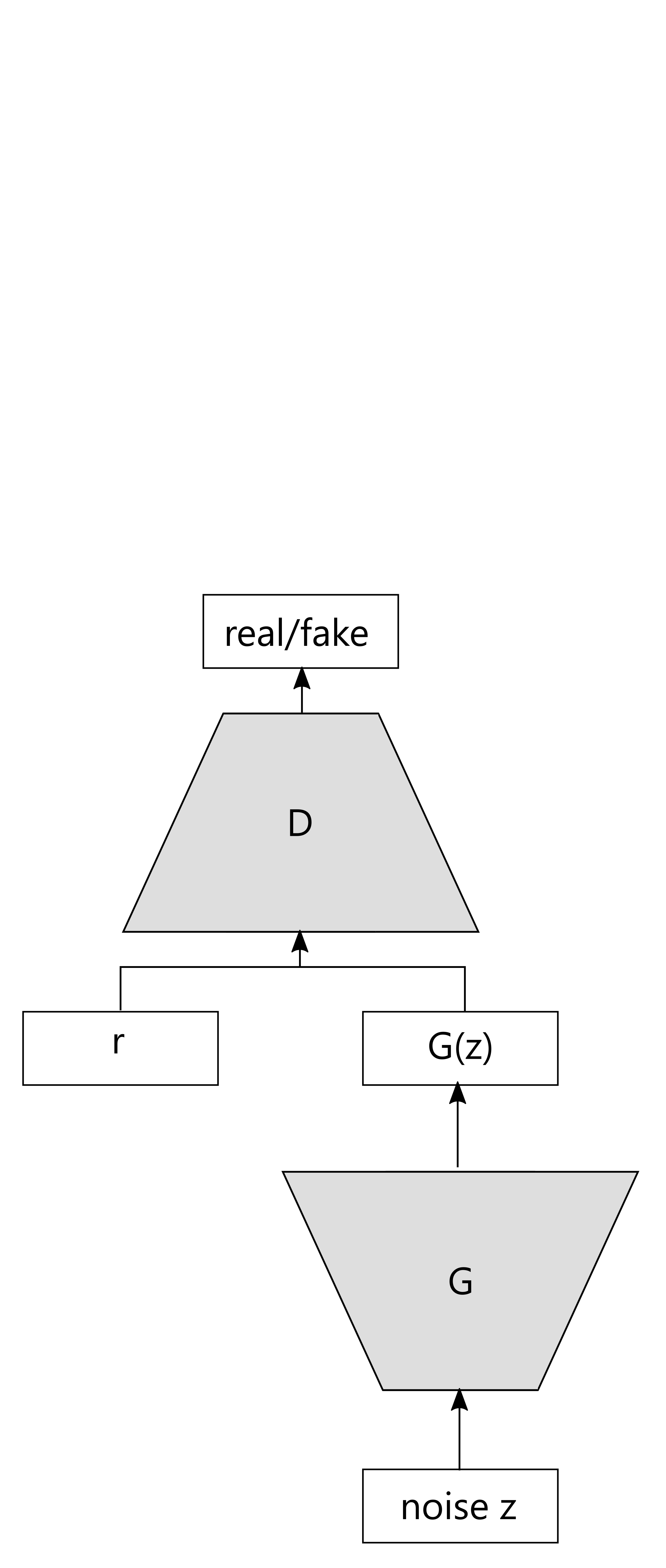}
\caption{DC-GAN\cite{radford2015unsupervised}}
\end{subfigure}%
\begin{subfigure}[t]{0.24\textwidth}
\centering
\includegraphics[width=\textwidth]{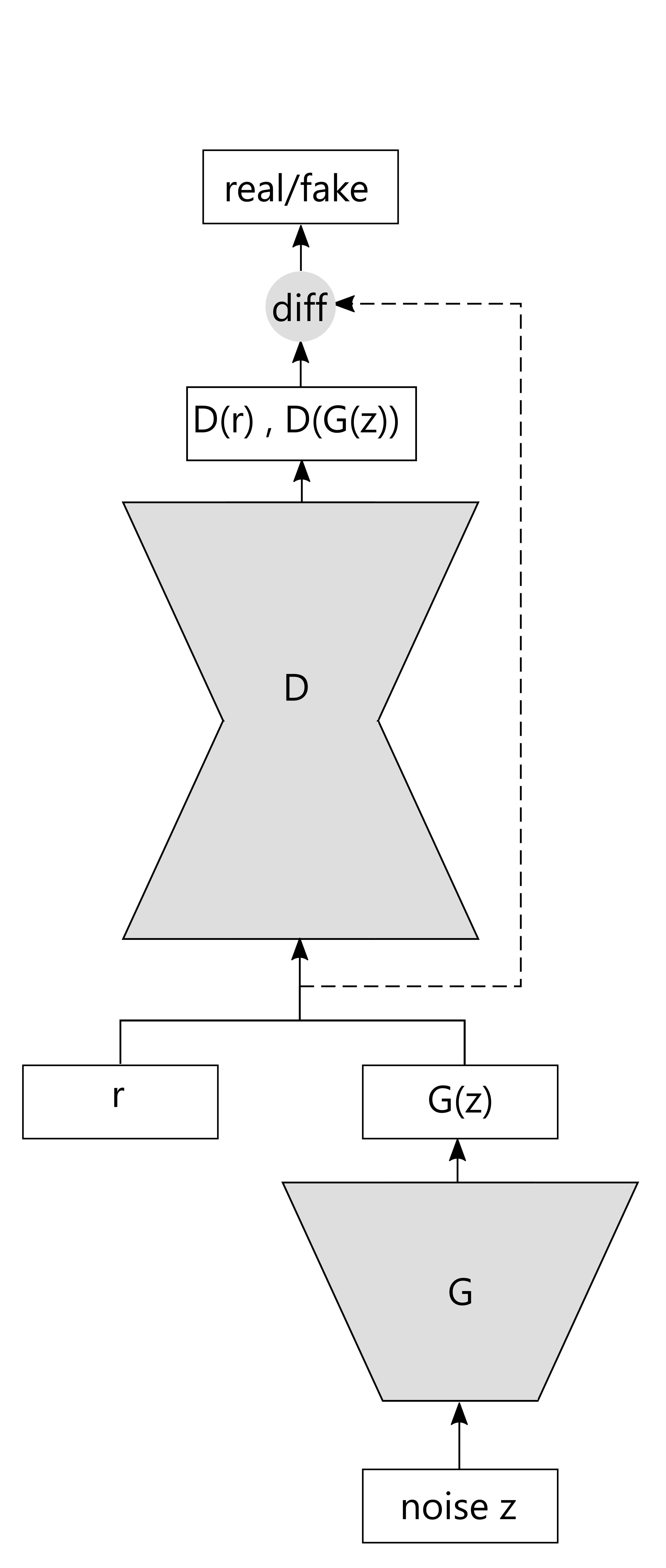}
\caption{BEGAN~\cite{berthelot2017began}}
\end{subfigure}%
\begin{subfigure}[t]{0.24\textwidth}
\centering
\includegraphics[width=\textwidth]{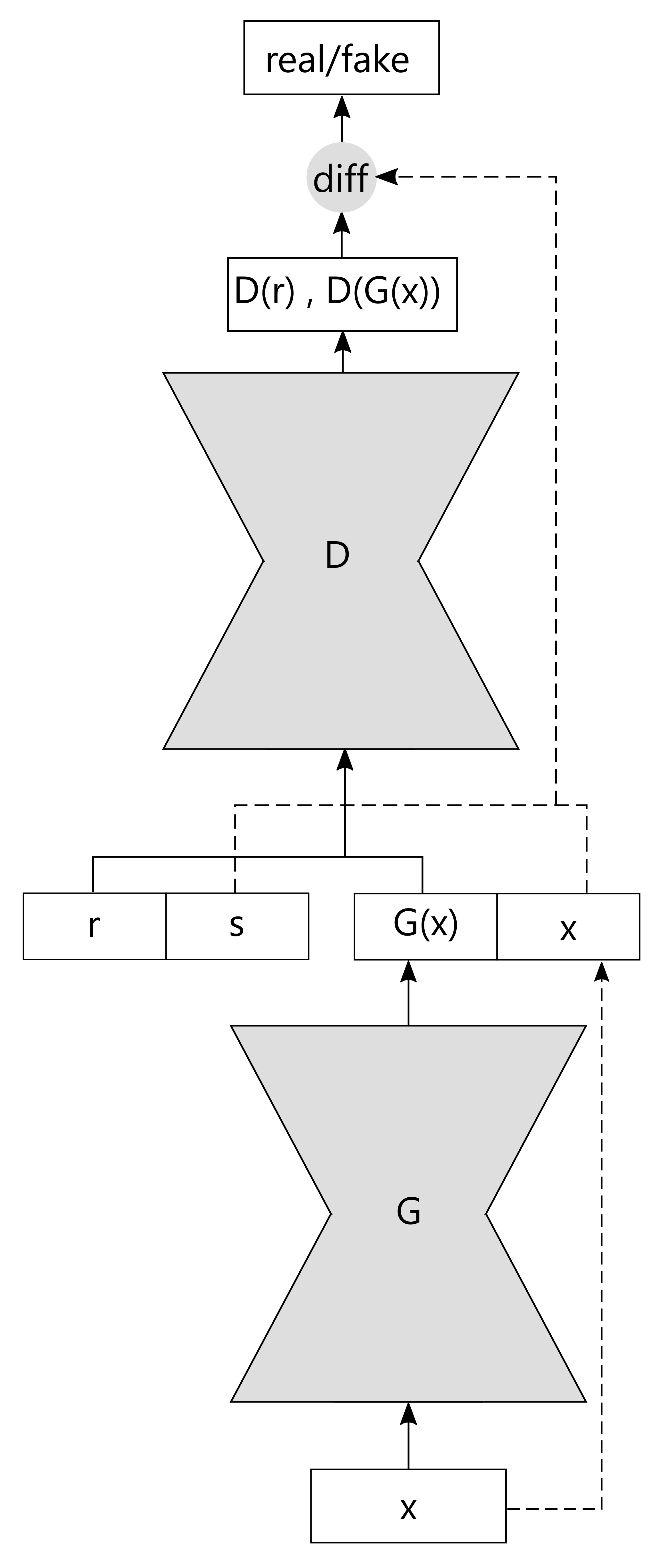}
\caption{Ours}
\end{subfigure}%
\begin{subfigure}[t]{0.24\textwidth}
\centering
\includegraphics[width=\textwidth]{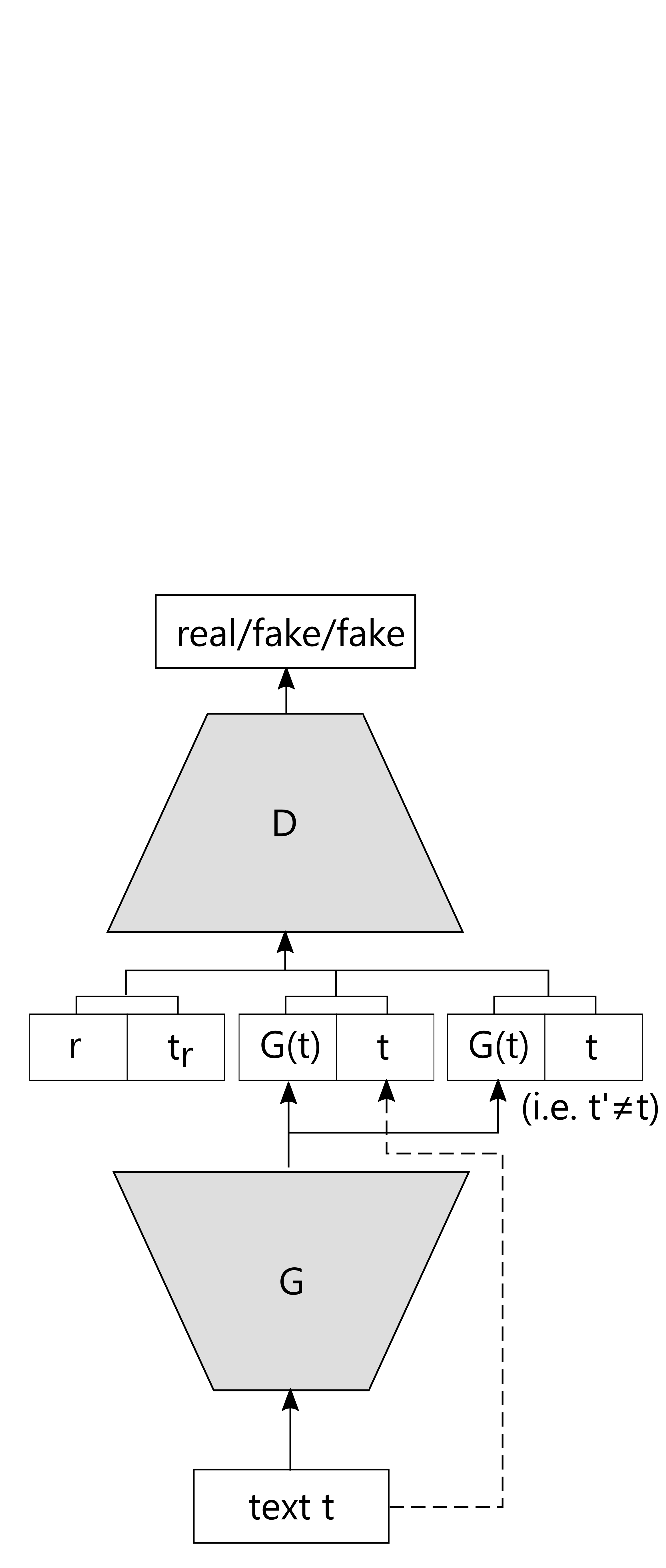}
\caption{GAN-CLS~\cite{reed2016generative}}
\end{subfigure}%
\caption{Comparison of our pair matching method to the related work. (a) In the traditional GAN approach, discriminator module align the distribution of real and synthetic images which is designed as a classification network. (b) BEGAN\cite{berthelot2017began} and many others showed that alignment of error distribution offers more stable training and better results. (c) We propose to utilize this autoencoder approach to align the distribution of pairs to encourage generated image to be a correct transformation to the realistic domain with a game between real and synthetic pairs. (d) An alternative to our method is to introduce wrongly labeled generated images to the discriminator to teach pair-wise matching.~\cite{reed2016generative} used such approach for text to images synthesis.}
\label{fig:methods}
\end{figure*}
Cycle consistency loss ensures bijective transitivity of functions $G$ and $G'$ which means generated image $G(x) \in \hat{R}$ should be transformed back to $x \in \hat{S}$. Convolutional networks are highly under-constrained and they are free to make any unintended changes as long as the cycle consistency is satisfied. Therefore, without additional supervision, it is not guaranteed to achieve the correct mapping that preserves shape, texture, expression, pose and lighting attributes of the face image from domains $S$ to $\hat{R}$ and $\hat{R}$ to $\hat{S}$. This problem is often addressed by introducing pixel-level penalization between input and output of the networks~\cite{zhu2017unpaired,shrivastava2016learning} which is sub-optimal for domain adaptation as it encourages to stay in the same domain.

To overcome this issue, we propose an additional pair-wise adversarial loss that assign $G'$ network an additional role as a pair-wise discriminator to supervise $G$ network. Given a set of paired synthetic and real images $(\mathcal{P_S},\mathcal{P_R})$, the discriminator loss is computed by BEGAN as follows:

\begin{align}
\mathcal{L}_{D_P} &= \mathds{E}_{s \in \mathcal{P_S},r \in \mathcal{P_R}}  \|s-G'(r)\|_1 - k^{D_P}_t \mathcal{L}_{cyc}
\end{align}

While $G'$ network is itself a generator network ($G':\hat{R} \rightarrow \hat{S}$) with a separate discriminator ($D_S$), we use it as a third pair-matching discriminator to supervise $G$ by means of distribution of paired correspondence of real and synthetic images. Thus while cycle-loss optimizes for biject correspondence, we expect resulting pairs of $(x \in S, G(x) \in \hat{R})$ to have similar correlation distribution as paired training data $(s \in \mathcal{P_S}, r \in \mathcal{P_R})$. Fig \ref{fig:methods} shows its relation to the previous related arts and comparison to an alternative which is matching aware discriminator with paired inputs for text to image synthesis as suggested by~\cite{reed2016generative}. Please notice that how BEGAN autoencoder architecture is utilized to align the distribution of pair of synthetic and real images with synthetic and generated images.

\subsection{Identity Preservation}
\label{sec:id_preservation}

Although identity information is provided by the 3DMM in shape and texture parameters, it may be lost to some extent by virtue of a non-linear transformation. Some studies~\cite{yin2017towards,tran2017disentangled} address this issue by employing identity labels of known subjects as additional supervision either with a pre-trained classification network or within the discriminator network. However, we intend to generate images of new identities sampled from 3DMM parameter space and their photorealistic images simply do not exist yet. Furthermore, training a new softmax layer and the rest of the framework simultaneously becomes a chicken-egg problem and results in failed training.

In order to preserve identity on the changing image space, we propose to adapt a set-based approach over a pre-trained face embedding network. We import the idea of pulling same-id samples as well as pushing close samples from different identities in the embedding space such that same-id images are gathered and distinct from other identities regardless of the quality of the images during the training. At the embedding layer of a pre-trained network $C$, generator network ($G$) is supervised by a combination of center~\cite{wen2016discriminative} and pushing loss~\cite{gecer2017learning}, which is also a simplified version of Magnet loss~\cite{rippel2015metric} formulation which is as following for a given mini-batch (M):
\begin{align}
\mathcal{L}_{C} = \mathds{E}_{x \in \mathcal{S},i_x \in \mathbb{N^+}} \sum_x^M -log\frac{\exp(\frac{1}{2\sigma^2}\|C(G(x))-c_{i_x}\|^2_2 -\eta)}{\sum_{j \neq i_x} \exp(\frac{1}{2\sigma^2}\|C(G(x))-c_j\|^2_2)}
\end{align}
where $i_x$ stands for the identity label of $x$ provided by 3DMM sampling. Margin term $\eta$ is set to 1 and the variance is computed by $\sigma = \frac{\sum_x^M\|C(G(x))-c_{i_x}\|^2_2}{M-1}$.

\begin{figure*}[t]
\includegraphics[width=\textwidth]{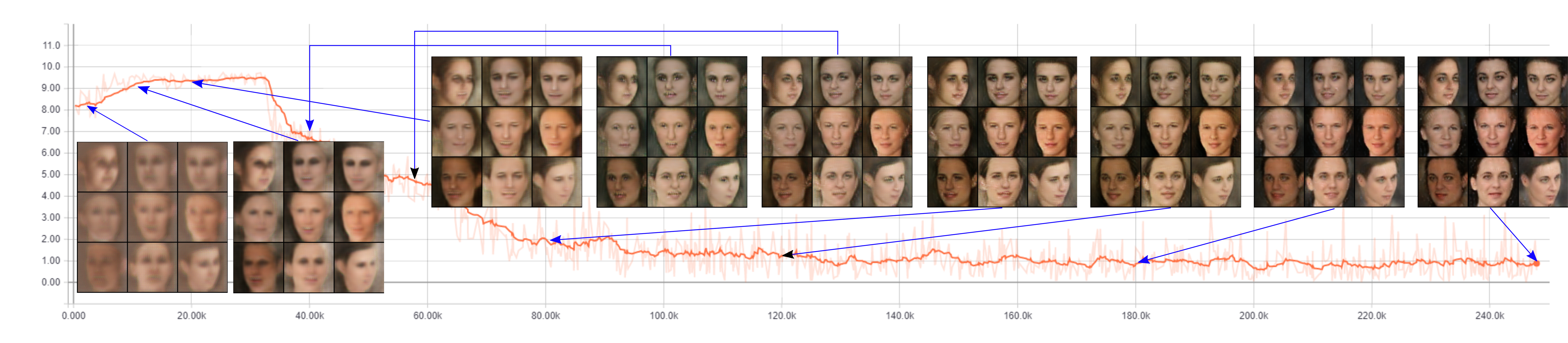}
\caption{Quality of 9 images of 3 identities (per row) during the training. Background plot shows the error by the proposed identity preservation layer over the iterations. Notice the changes on the level of fine-details on the faces which is the main motivation of using set-based identity preservation.}
\label{fig:c_loss}
\end{figure*}

While the quality of images is improved during the training, their projection on the embedding space is shifting. In order to adapt to those changes, we update identity centroids ($c_j$) with a momentum of $\beta=0.95$ when new images of id $j$ is available. Following~\cite{wen2016discriminative}, for a given $x$, moving average of a identity centroid  is calculated by $c_j^{t+1} = c_j^t - \beta\delta(i_x=j)(c_j^t-C(G(x))) $ where $\delta(condition)=1$, if the condition is satisfied and $\delta(condition)=0$ if not. Centroids ($c_j$) are initialized with zero and after few iterations, they converge to embedding centers and then continue updating to adapt to the changes caused by the simultaneous training of $G$. Fig. \ref{fig:c_loss} shows quality of 9 images of 3 identities over training iterations. Please notice the difference of the images after convergence with the images at the beginning of the training which Softmax layer might converge and fail to supervise for the forthcoming images in later iterations.

\subsection*{Full Objective}
Overall, the framework is optimized by the following updates simultaneously:
\begin{align}
\theta_G &= \mathop{\argmin}_{\theta_G} \mathcal{L}_G + \lambda_{cyc}\mathcal{L}_{cyc} + \lambda_C\mathcal{L}_C
\label{eq:full1}\\
\theta_{G'} &= \mathop{\argmin}_{\theta_{G'}} \mathcal{L}_{G'} + \lambda_{cyc}\mathcal{L}_{cyc} + \lambda_{D_P}\mathcal{L}_{D_P}
\label{eq:full2}\\
\theta_{D_R},\theta_{D_S} &=\mathop{\argmin}_{\theta_{D_R},\theta_{D_S}} \mathcal{L}_{D_R} + \mathcal{L}_{D_S}
\label{eq:full3}
\end{align}
where $\lambda$ parameters balance the contribution of different modules. The selection of those parameters is discussed in the next section.
\section{Implementation Details}

\paragraph{Network Architecture:}
For the generator networks ($G$ and $G'$), we use a shallow ResNet architecture as in~\cite{johnson2016perceptual} which supplies smooth transition without changing the global structure because of its limited capacity with only 3 residual blocks. In order to benefit from 3DMM images fully, we also add skip connections to the network $G$. Additionally, we add dropout layers after each block in the forward pass with a 0.9 keep rate in order to introduce some noise that could be caused by uncontrolled environmental changes. 

We construct the discriminator networks ($D_R$ and $D_S$) as autoencoders trained by boundary equilibrium adversarial learning with Wasserstein distance as proposed by~\cite{berthelot2017began}. The classification network $C$, is a shallow FaceNet architecture~\cite{schroff2015facenet}, more specifically we use NN4 network with an input size of $96 \times 96$ where we randomly crop, rotate and flip generated images $G(x)$ which are in size of $108 \times 108$.

\paragraph{Data:} Our framework needs a large amount of real and synthetic of face images. For real face images, we use CASIA-Web Face Dataset~\cite{yi2014learning} that consists of \~500K face images of \~10K individuals.


Please recall that the proposed method trains the $G'$ network as a discriminator ($D_P$) with a small number of paired examples of real and synthetic images. For that, we use a combination of 300W-3D and AFLW2000-3D datasets as our paired training set~\cite{zhu2016face} which consist of 5K real images with their corresponding 3DMM parameter annotations. We render synthetic images by those latent parameters and pair them with matching the real images. This dataset relatively small compared to the ones used by fully supervised transformation GANs (i.e. Amazon Handbag dataset used by~\cite{isola2016image} contains 137K bag images)

We randomly sample 500K face images of 10K identities as our synthetic data set using Large Scale Face Model (LSFM)~\cite{booth20163d} and Face Warehouse model for expressions~\cite{cao2014facewarehouse}. While shape and texture parameters of new identities are sampled to be under Gaussian distribution of the original model, expression, lighting and pose parameters are sampled with the same Gaussian distribution as synthetic samples of 300W-3D and AFLW2000-3D. For our experiments, we align the faces using MTCNN~\cite{zhang2016joint} and centre crop them to the size of $108 \times 108 \times 3$ pixels. 
\paragraph{Training Details:}
We train all the components of our framework together from scratch except the classification network $C$ which is pre-trained by using a subset of Oxford VGG Face Dataset~\cite{parkhi2015deep}. The whole framework takes about 70 hours to converge on a Nvidia GTX 1080TI GPU for 248K iterations with batch size of 16. We start with a learning rate of $8\times10^{-5}$ with ADAM solver~\cite{kingma2014adam} and halve it at after 128Kth, 192Kth, 224Kth, 240Kth, 244Kth, 246Kth and 247Kth iterations.

As shown in Eqn. \ref{eq:full1}, \ref{eq:full2}, $\lambda$ is a balancing factor which controls the contribution of each optimization. We set $\lambda_{cyc} = 0.5$, $\lambda_{D_P} = 0.5$, $\lambda_{C} = 0.001$ to balance between realism, cycle-consistency, identity preservation and the supervision by the paired data. We also add identity loss ($\mathcal{L}_{id} = \|x - G(x)\|$) as suggested by~\cite{zhu2017unpaired} to regularize the training with a balancing term $\lambda_{id}=0.1$. During the training, we keep track of moving averages of the network parameters to generate images.

As side notes, in our experiments, we observed that it is beneficial to keep non-adversarial signals weak to avoid mode collapse. We also observed that the approach of keeping the history of refined images proposed by~\cite{bousmalis2016unsupervised} breaks adversarial training in our case due to the auto-encoder discriminators. 

\section{Results and Discussions}
\label{sec:experiments}

\begin{figure*}[t]
\includegraphics[width=1.0\textwidth]{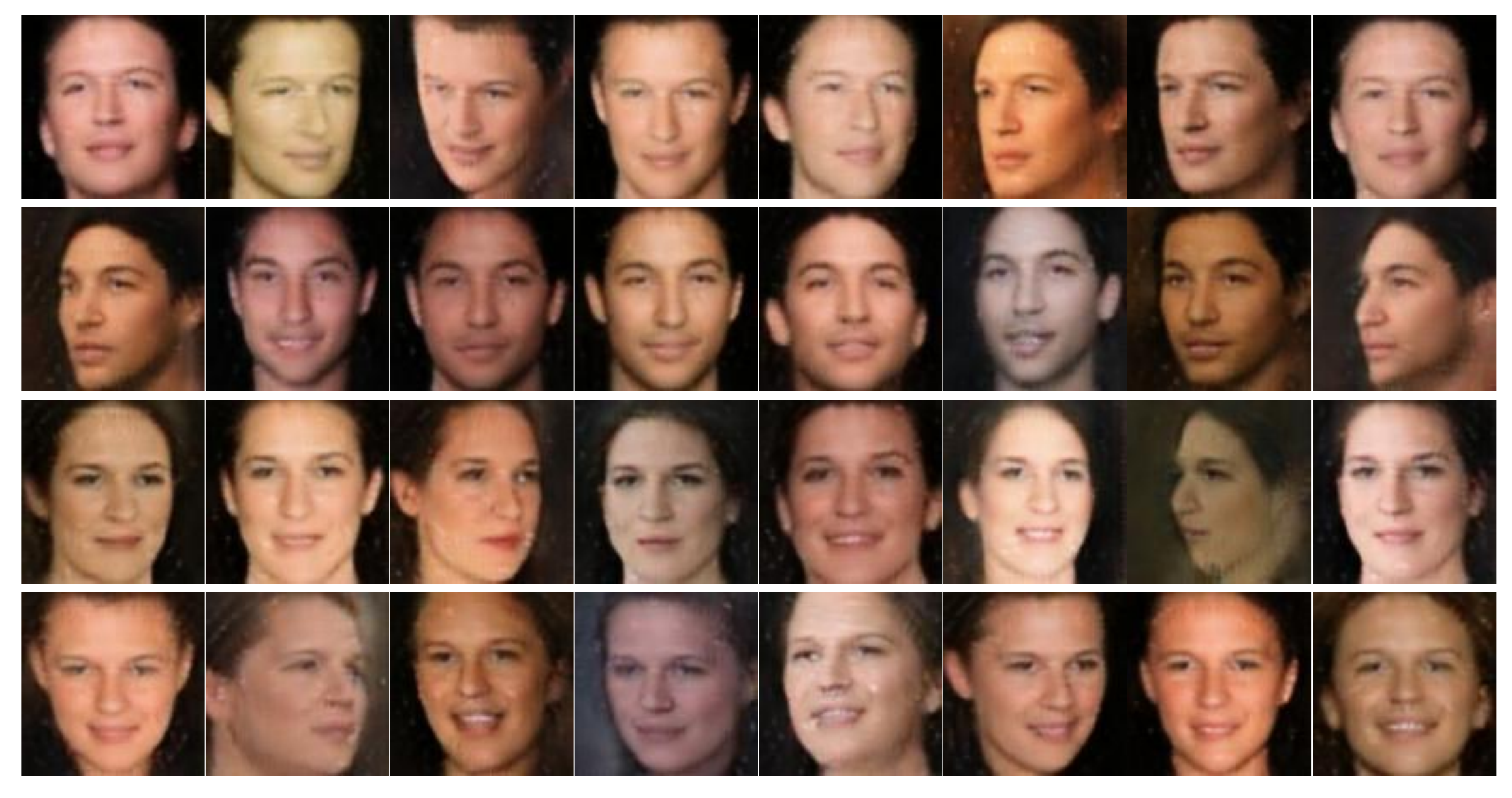}
\caption{Random samples from GANFaces dataset. Each row belongs to same identity. Notice the variation in pose, expression and lighting.}
\label{fig:randomsamples}
\end{figure*}

In this section, we show qualitative and quantitative results of the proposed framework. We also discuss and show the contribution of each module (i.e. $\mathcal{L}_{cyc}$, $D_P$, $C$) with an ablation study in the supplementary materials. For the experiments,  we generate 500,000 images of 10,000 different identities with variations on expression, lighting and poses. We name this synthetic dataset \textbf{GANFaces}. Please see Fig.\ref{fig:randomsamples} for random samples from the dataset. The dataset, training code, pre-trained models and face recognition experiments can be viewed at~\underline{\url{https://github.com/barisgecer/facegan}}.

\subsection{Visually Plausible 3DMM Generation}
\begin{figure*}[t]
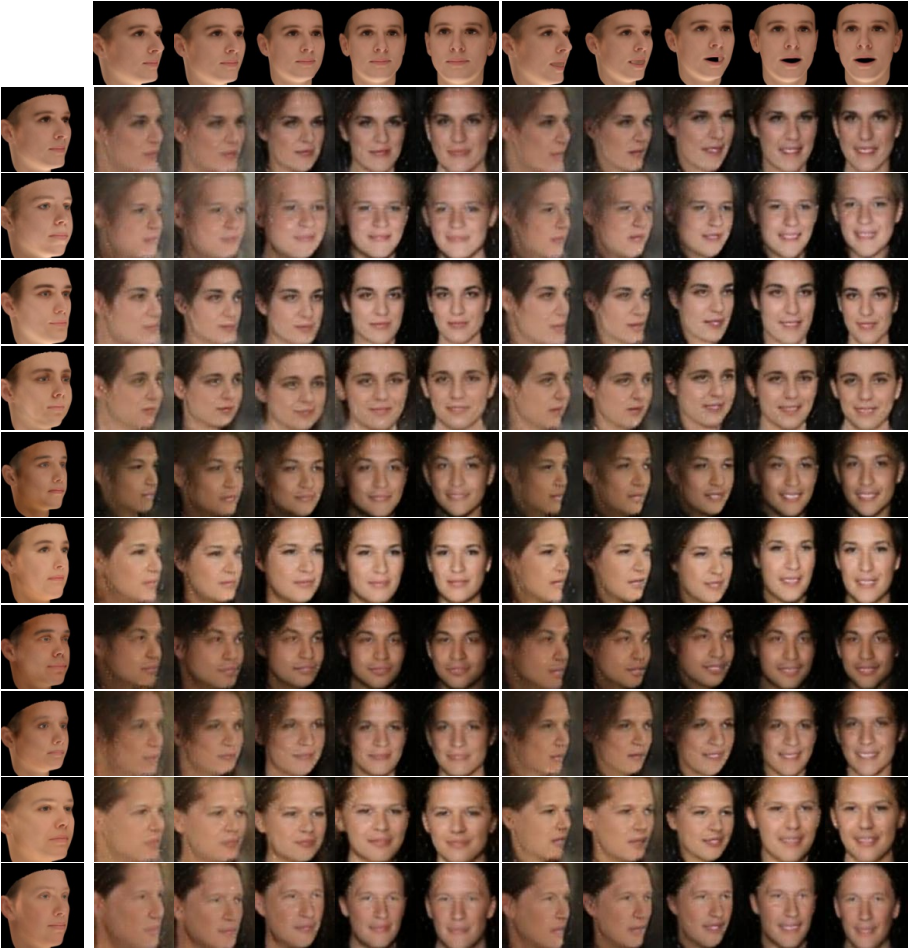

\hspace{1.095cm}
\foreach \e in {1,...,2}{
\foreach \x in {03,05,07,09,11}{\hspace{-0.02cm}\includegraphics[width=0.09\textwidth]{figures/eccv-108/00006/\x_\e.jpg}}\hspace{-0.005cm}}\\
\foreach \i in {06,11,13,18,19,26,27,30,33,34}{\includegraphics[width=0.09\textwidth]{figures/eccv-108/000\i/04_1.jpg}
\foreach \e in {1,...,2}{
\foreach \x in {03,05,07,09,11}{
\hspace{-0.1cm}\includegraphics[width=0.09\textwidth]{figures/gen-eccv/000\i/\x_\e.jpg}}}\\
}
\caption{Images generated by the proposed approach conditioned with identity variation in vertical axis, normalized and mouth open expression in left and right blocks and pose variation in horizontal axis. Images in this figure are not included in the training }
\label{fig:variation}
\end{figure*}

One of the main goals of this work is to generate the face images guided by the attributes of synthetic input images \ie shape, expression, lighting, and poses. We can see from the Fig.~\ref{fig:variation} that our model is capable of generating photorealistic images preserving the attributes conditioned by the synthetic input images. 
In the Figure, top row shows the variations of pose and expression on input synthetic faces and the left column shows the input synthetic faces of different identities. And, the rest of the images are the images generated by our model conditioned on the corresponding attributes from top row and left column. 
We can clearly see that the conditioned attributes are preserved on the images generated by our model. We can also observe that fine-grained attributes such as shapes of chin, nose and eyes are also retained on the images generated by our model. In case of extreme poses, the quality of the image generated by our model becomes less sharp as the CASIA-WebFace dataset, which we used to learn the parameters of discriminator network $D_R$, lacks sufficient number of examples with extreme poses. 


\subsection{The Added Realism and Identity Preservation}
\begin{figure*}[t]
\includegraphics[width=0.33\textwidth, trim= 80 260 120 260, clip]{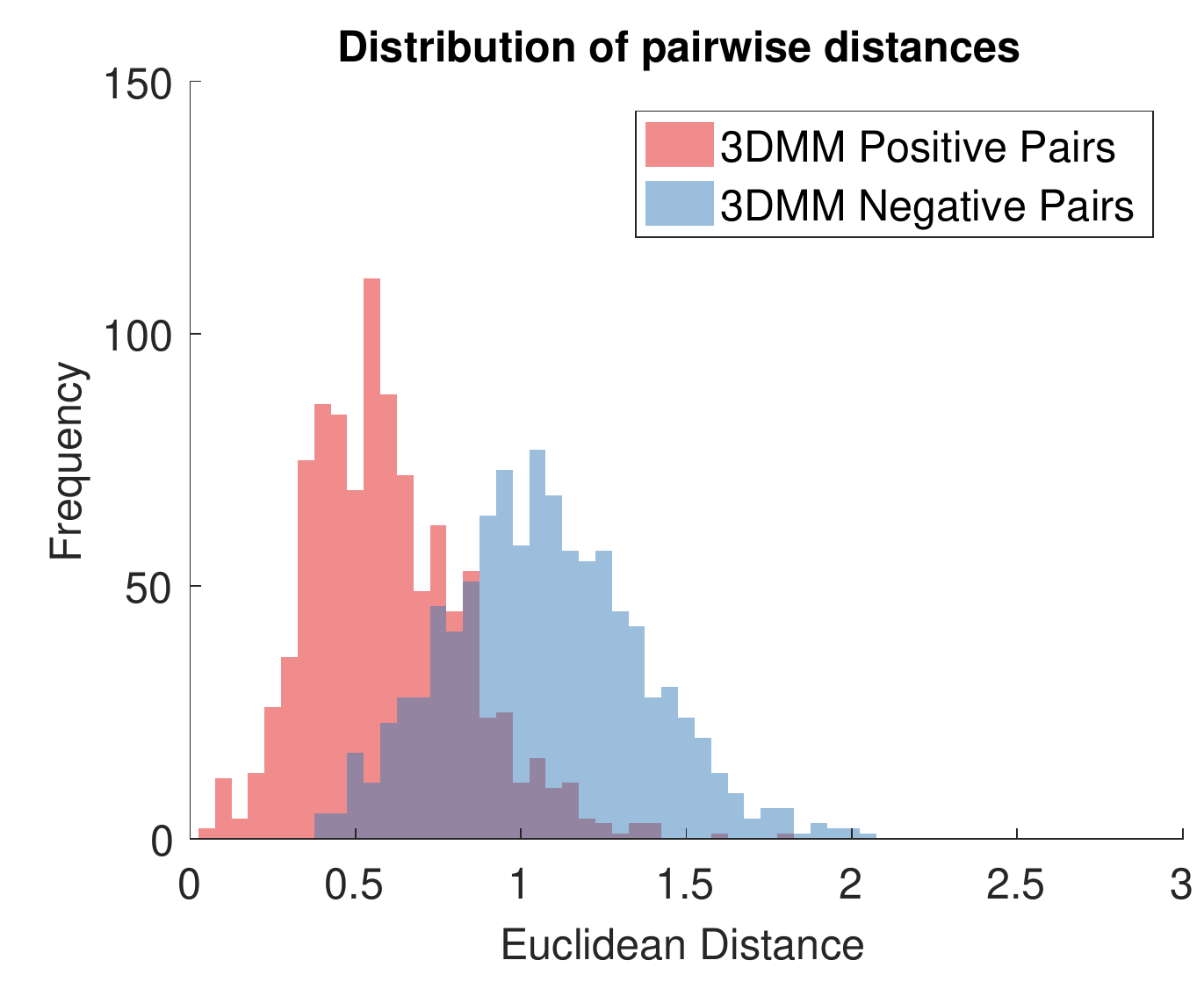}
\includegraphics[width=0.33\textwidth, trim= 80 260 120 260, clip]{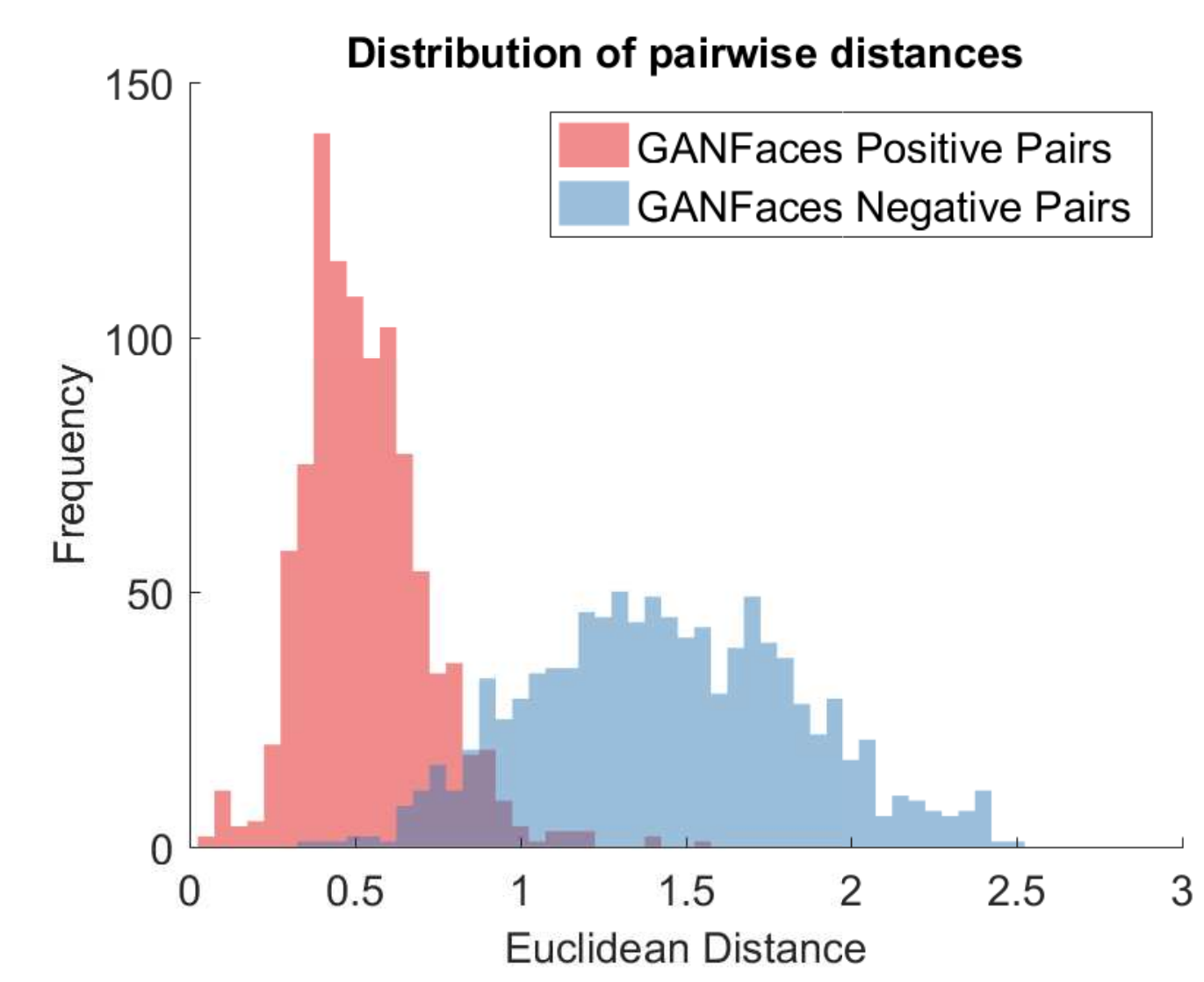}
\includegraphics[width=0.33\textwidth, trim= 80 260 120 260, clip]{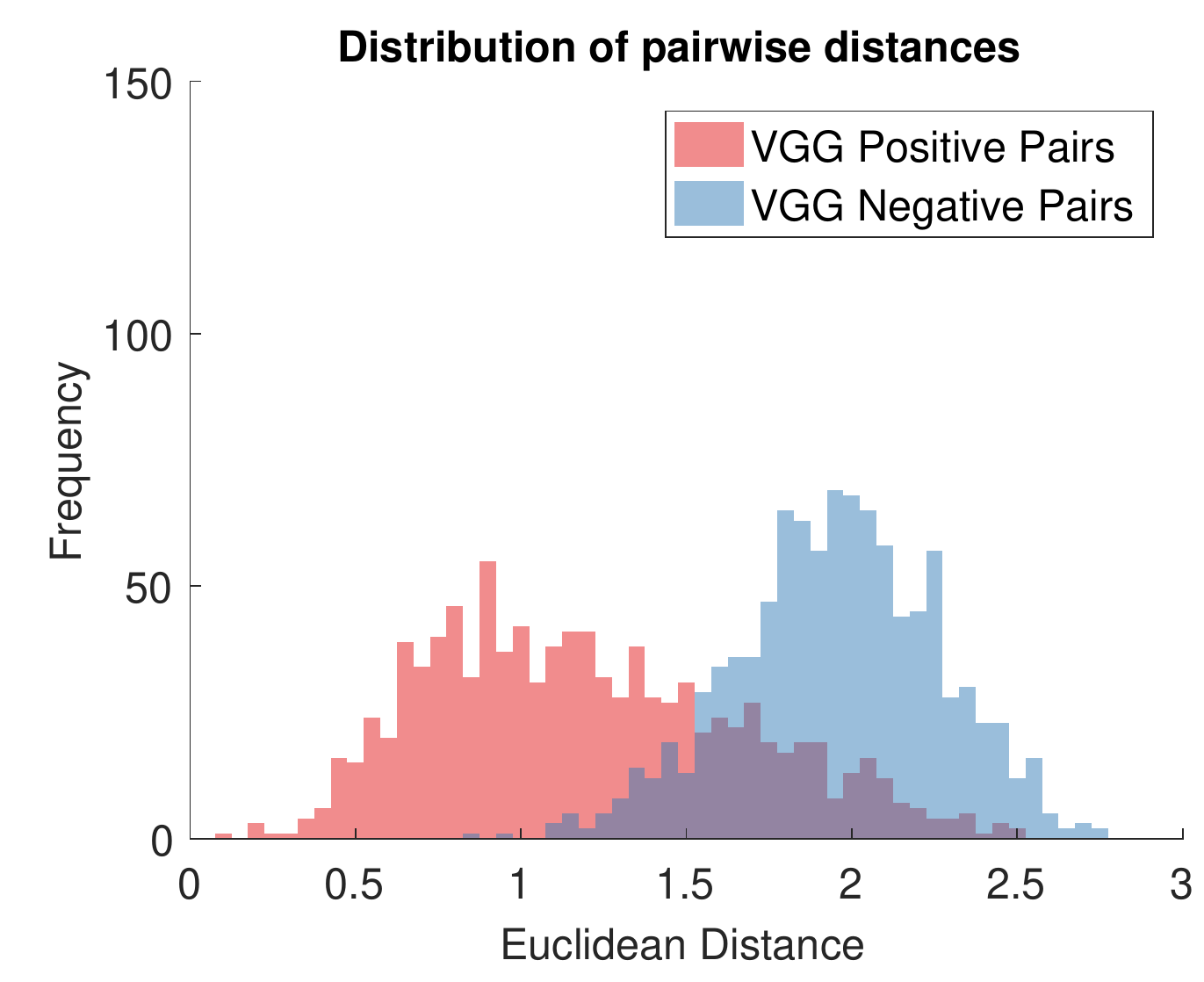}
\caption{Distances of 1000 positive and 1000 negative pairs from three different datasets (GANFaces, 3DMM synthetic images, Oxford VGG) embedded on a NN4 network that is trained with CASIA Face dataset}
\label{fig:distances}
\end{figure*}

In order to show that synthetic images are effectively transformed to the realistic domain with preserving identities, we perform a face verification experiments on GANFaces dataset. We took pre-trained face-recognition CNN network, namely FaceNet NN4 architecture~\cite{schroff2015facenet} trained on CASIA-WebFace~\cite{yi2014learning} to compute the features of the face images. The verification performance of the network on LFW is $\%95.6$ accuracy and $\% 95.5$ 1-EER which shows that the model is well optimized for in-the-wild face verification. We created 1000 similar (belonging to same identity) and 1000 dis-similar(belonging to different identities) face image pairs from GANFaces. Similarly, we also generated the same number of similar and dis-similar face images pairs from VGG face dataset
\cite{parkhi2015deep} and the synthetic 3DMM rendered faces dataset. Fig.~\ref{fig:distances} shows histogram of euclidean distances between similar and dis-similar images measured in the embedding space for the three datasets. The addition of realism and preservation of identities of the GANFaces can be seen from the comparison of its distribution to the 3DMM synthetic dataset's distribution. As the images become more realistic, they become better separable in the pre-trained embedding space. We also observe that the separation of positive and negative pairs of GANFace's faces are better than that of VGG faces pairs. The probable reason of VGG does not having better separation than GANFaces is due to noisy face labels and this is indicated on its original study~\cite{parkhi2015deep}.


\subsection{Face Recognition with GANFaces dataset}
We augmented GANFaces with real face dataset \ie VGG Faces~\cite{parkhi2015deep} and train VGG19~\cite{Simonyan14c} network  and tested performance on two challenging datasets: Labeled Faces in the Wild (LFW)~\cite{huang2007labeled} and IJB-A~\cite{klare2015pushing}. We restrict ourselves from limited access to full access of real face dataset and train deep network on different combination of real and GANFaces. Following~\cite{masi2016we}, we use a pre-trained VGGNet by~\cite{Simonyan14c} with 19 layers trained on ImageNet dataset~\cite{russakovsky2015imagenet} and took these parameters as initial parameters. We train the network with different portion of Oxford VGG Face dataset~\cite{parkhi2015deep} augmented with the GANFaces dataset. We remove the last layer of deep VGGNet and add two soft-max layers to the previous layer, one for each of the datasets. Learning rate is set to 0.1 for the soft-max layers and 0.01 to the pre-trained layers with ADAM optimizer. Also we halve the gradient coming from GANFaces soft-max. We decrease the learning rate exponentially and train for 80,000 iterations where all of our models are well converged without overfitting. For a given input size of $108 \times 108$, we randomly crop and flip $96 \times 96$ patches and overall training takes around 9 hours on a NVIDIA 1080TI GPU.

\begin{figure*}[t]
\includegraphics[width=0.33\textwidth, trim= 80 260 110 260, clip]{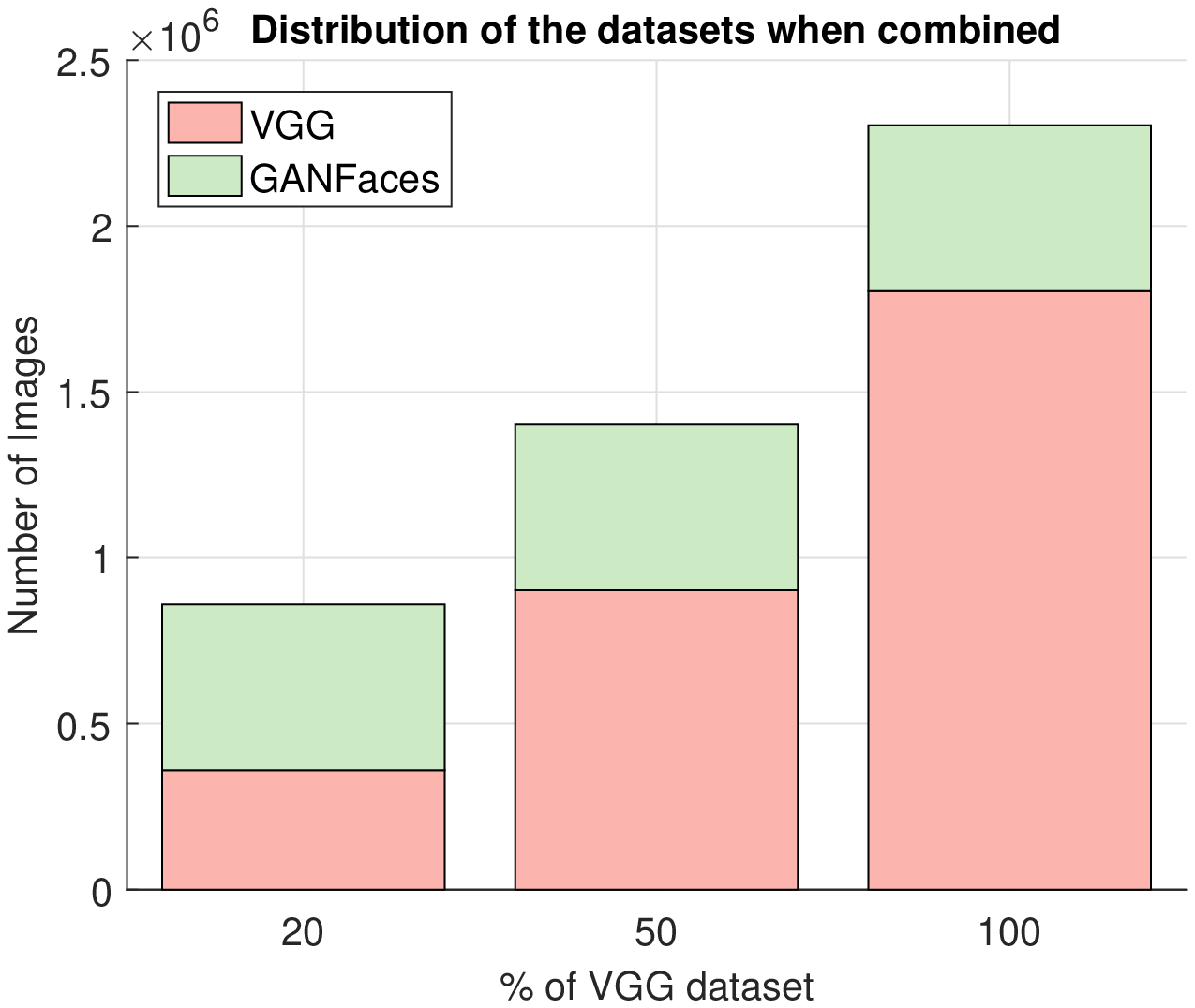}
\includegraphics[width=0.33\textwidth, trim= 80 260 110 260, clip]{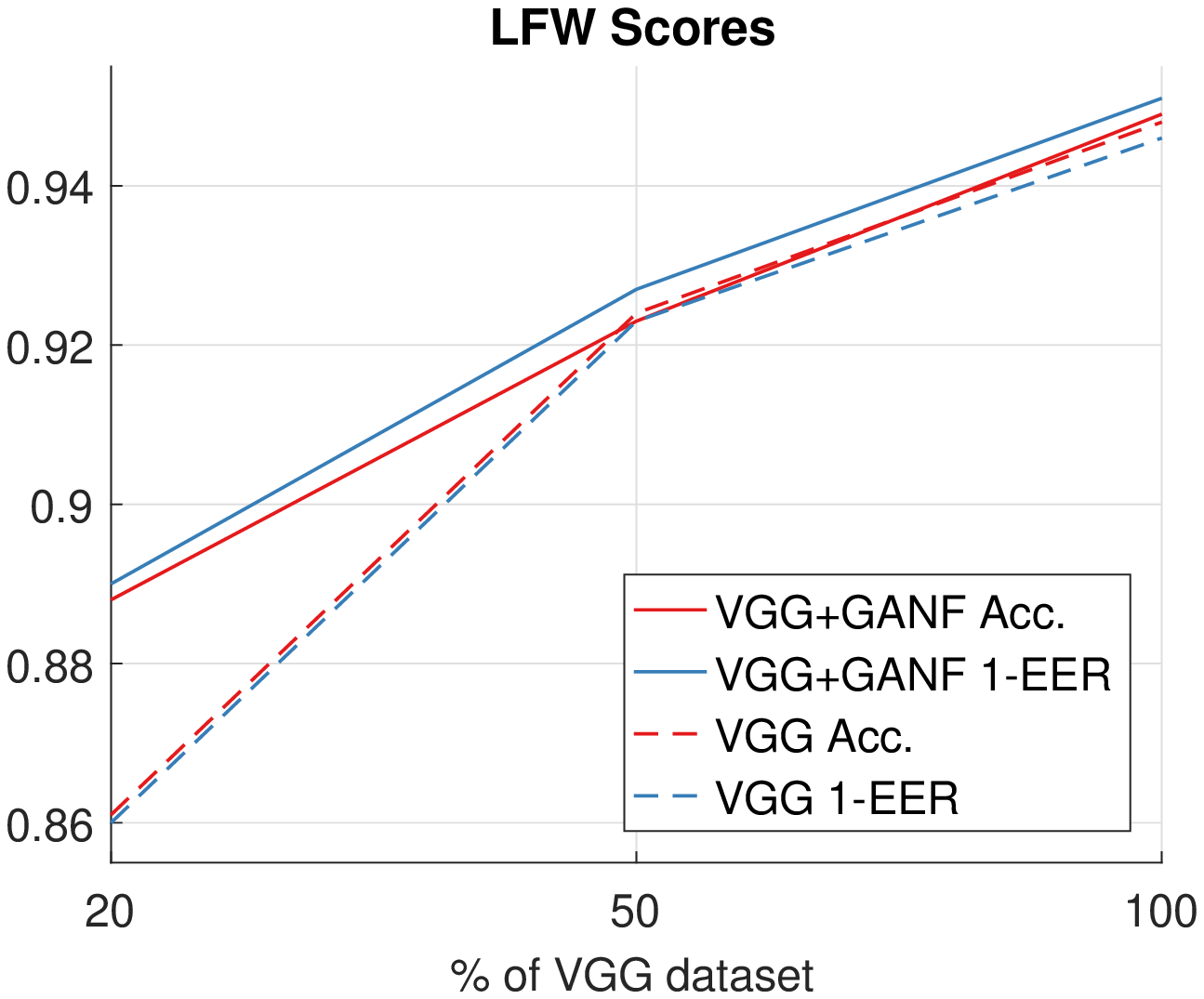}
\includegraphics[width=0.33\textwidth, trim= 80 260 110 260, clip]{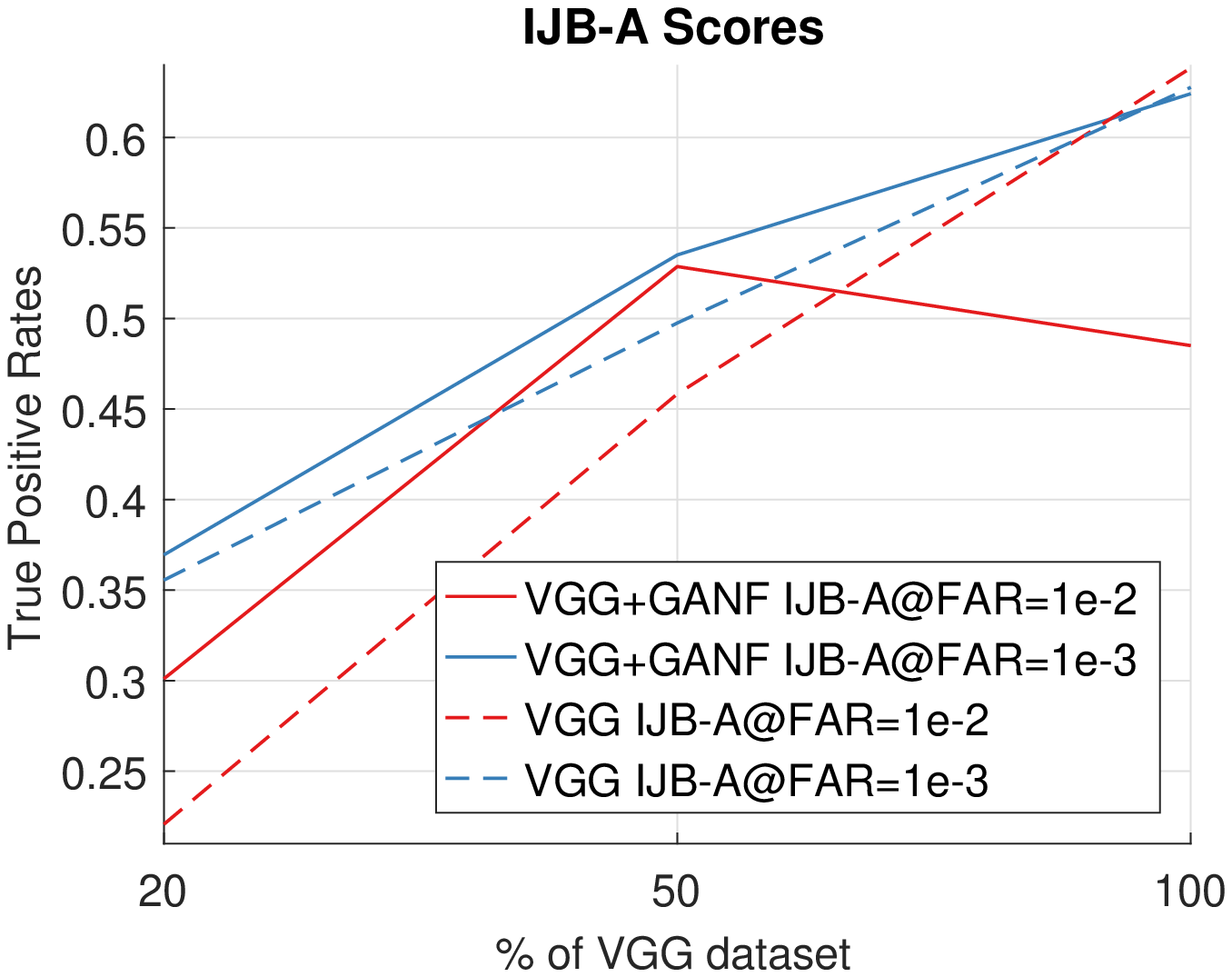}
\caption{Face recognition benchmark experiments. (Left) Number of images used from the two datasets in the experiments. Total number of images of VGG Data set is 1.8M since some images were removed from the URL (Middle) Performances on the LFW dataset with and without GANFaces dataset. (Right) True Positive Rates on IJB-A verification task  with and without GANFaces dataset.}
\label{fig:results}
\end{figure*}

We train 6 models with $\%20$, $\%50$ and $\%100$ of the VGG Face dataset with and without the augmentation of GANFaces. We evaluate the models on LFW and IJB-A datasets and the benchmark scores is improved with the usage of GANFaces dataset even though low resolution images. The contribution of GANFaces increase inversely proportional to the number of images included from VGG dataset which indicates more synthetic images might improve the results even further. Further details can be seen in Fig. \ref{fig:results}.

We compare our best model with full VGG dataset and GANFaces to the other state of the art methods. Despite the very low resolution compared to the others, GANFaces was able to improve our baseline to the numbers comparable to the state-of-the-arts. Please note that generative methods such as~\cite{masi2016we,yin2017towards}, do generation (i.e. pose augmentation and normalization) in the test time where we use only given test images. Together with low resolution, this makes our models more efficient at test time. Given that we only generated 500K images show that the accuracy can be boosted even further by generating more (i.e. 5 times larger from the real set as~\cite{masi2016we}).

\begin{table*}
\begin{center}
\begin{tabular}{|c|c|c|c|c|c|c|c|c|}
\hline
Method & Real & Synth & Test time Synth & Image size & Acc. ($\%$) & 100$\%$ - EER \\
\hline
FaceNet~\cite{schroff2015facenet} & 200M & - & No & 256$\times$256 & 98.87 & - \\
VGG Face~\cite{parkhi2015deep} & 2.6M & - & No & 256$\times$256 & 98.95 & 99.13 \\
\hline
Masi \etal~\cite{masi2016we} & 495K & 2.4M & Yes & 256$\times$256 & 98.06 & 98.00\\
Yin  \etal~\cite{yi2013towards} & 495K & 495K & Yes & 256$\times$256 & 96.42& - \\
\hline
\hline
VGG($\%100$) & 1.8M & - & No & 108$\times$108 & 94.8 & 94.6 \\
VGG($\%100$) + GANFaces & 1.8M & 500K & No & 108$\times$108 & \textbf{94.9} & \textbf{95.1}\\
\hline
\end{tabular}
\end{center}
\caption{Comparison with state-of-the-art studies on LFW performances}
\label{tab:userstudy1}
\end{table*}

\section{Conclusions}
\label{sec:conclusions}


In this paper, we propose a novel end-to-end semi-supervised adversarial training framework to generate photorealistic faces of new identities with wide ranges of poses, expressions, and illuminations from 3DMM rendered faces. Our extensive qualitative and quantitative experiments show that the generated images are realistic and identity preserving.

We generated a dataset of 500,000 face images and combined it with a real face image dataset to train a face recognition CNN and improve the performances in recognition and verification tasks. Despite the limited the number of images generated, they were still enough to improve recognition rates. In the future, we plan to generate millions of high resolution images of thousands of new identities to boost the state-of-the-art face recognition.

\section*{Acknowledgments}
This work was supported by the EPSRC Programme
Grant `FACER2VM' \\(EP/N007743/1). Baris Gecer is funded by the Turkish Ministry of National Education. This study is morally motivated to improve face recognition to help prediction of genetic disorders visible on human face in earlier stages.


{\small
\bibliographystyle{ieee}
\bibliography{biblio}
}

\newpage
\title{Semi-supervised Adversarial Learning to Generate Photorealistic Face Images of New Identities from 3D Morphable Model: Supplementary Material} 

\titlerunning{Semi-supervised Adv. Learning to Generate Face Images of New Ids from 3DMM}

\authorrunning{B. Gecer,  B. Bhattarai, J. Kittler, and  T.K. Kim} 

\author{
Baris Gecer\affmark[1], Binod Bhattarai\affmark[1], Josef Kittler\affmark[2], and Tae-Kyun Kim\affmark[1]
}

\institute{\affaddr{\affmark[1]Department of Electrical and Electronic Engineering,
Imperial College London, UK}\\
\affaddr{\affmark[2]Centre for Vision, Speech and Signal Processing, University of Surrey, UK}\\
\email{ \{b.gecer,b.bhattarai,tk.kim\}@imperial.ac.uk, j.kittler@surrey.ac.uk}}

\maketitle

\section{Ablation Study}
\subsection{Quantitative Results}
We investigate the contributions of three main components of our framework by an ablation study. Namely, identity preservation module ($\mathcal{L}_C$), adversarial pair matching $\mathcal{L}_{D_P}$ and cycle consistency loss $\mathcal{L}_{cyc}$\footnote{Here we do not investigate the contribution of the discriminators as their effect is shown by many other studies.}. We train our framework from scratch in the same way as explained in the paper by removing each of these modules separately (\ie for VGG ($\%$50) version). In table \ref{tab:userstudy1}, we show the contribution of each module and compare them to the whole framework as a baseline and to the performance of a model trained by only half of the VGG dataset.

\begin{table*}
\begin{center}
\begin{tabular}{|c|c|c|c|c|c|c|c|c|}
\hline
Method & IJB-A Ver. @FAR=0.01 &  IJB-A Ver. @FAR=0.001 \\
\hline
Ours without $\mathcal{L}_C$&0.50532 $\pm$ 0.00433 & 0.17636 $\pm$ 0.00611\\
Ours without $\mathcal{L}_{D_P}$&0.48034 $\pm$ 0.00402 & 0.15300$\pm$ 0.00348\\
Ours without $\mathcal{L}_{cyc}$&0.49701 $\pm$ 0.00558 & 0.18341$\pm$ 0.00670\\
\hline
\hline
VGG ($\%$50)&0.49751 $\pm$ 0.00484 & 0.17580 $\pm$ 0.00557 \\
Ours (VGG($\%$50)+GANFaces)& \textbf{0.53507 $\pm$ 0.00575} & \textbf{0.18768 $\pm$ 0.00388}\\
\hline
\end{tabular}
\end{center}
\caption{Quantitative ablation study. Each of the modules removed from the proposed framework and the performance of generated images are measured on IJB-A verification task}
\label{tab:userstudy1}
\end{table*}

\subsection{Qualitative Results}
\begin{figure*}
\foreach \m / \mi in {c/a,s/b,f/c,d/d,r/e}{(\mi)\hspace{-0.2cm}
\foreach \i in {5,6,7,3}{
\foreach \x in {a,b,c}{
\includegraphics[width=0.08\textwidth]{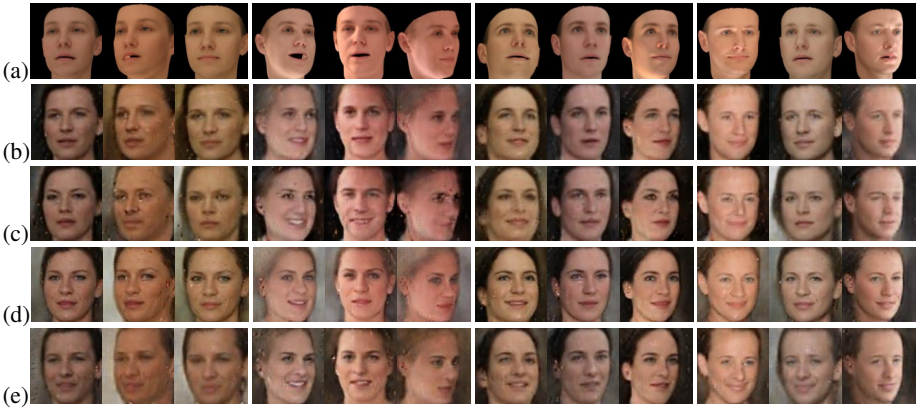}\hspace{-0.08cm}}}\\
}
\caption{Columns: divided into blocks of 3 images from the same identity. Rows: (a) 3DMM synthetic images. (b) Generated images by the framework (Ours). (c) Ours without $\mathcal{L}_C$. (d) Ours without $\mathcal{L}_{D_P}$. (e) Ours without $\mathcal{L}_{cyc}$.}
\label{fig:ablation_qul}
\end{figure*}

Fig. \ref{fig:ablation_qul} shows visual comparisons between the proposed framework and its versions without each of its components. For the framework and its three variants, we show generated images for 12 3DMM input images of 4 different identities with random illumination, pose and expression variations. We evaluate the quality of the images by identity preservation, the visual plausibility and diversity (\ie avoiding mode collapse). Regarding these criteria, our framework clearly generates better images than all of its variants. Without 
$\mathcal{L}_C$ (Fig.\ref{fig:ablation_qul}(c)), namely identity preservation module, the framework \textit{forgets} identity information throughout the network as there is no direct signal to encourage identity preservation. Please notice the identity consistency of our framework (Fig.\ref{fig:ablation_qul}(b)) compared to (Fig.\ref{fig:ablation_qul}(c)) in the details of faces (\ie. shape of nose, eyes, month, eyebrows and their relative distances) or simply by visual gender test. Without adversarial pair matching mechanism $\mathcal{L}_{D_P}$ (Fig.\ref{fig:ablation_qul}(d)), we observe local mode collapse across different identities such as shape of nose and eyes in the figure seems to be similar compared to Fig.\ref{fig:ablation_qul}(b). This mode collapse is also verified by the quantitative experiments (Table. \ref{tab:userstudy1}). Cycle consistency loss $\mathcal{L}_{cyc}$ (Fig.\ref{fig:ablation_qul}(e)) helps to retain the initial shape given by 3DMM and improves the overall image quality. We also observe noise reduction in the generated images due to the additional supervision provided by all the modules. 

\section{Identity and Illumination Interpolations}
\begin{figure*}[t]
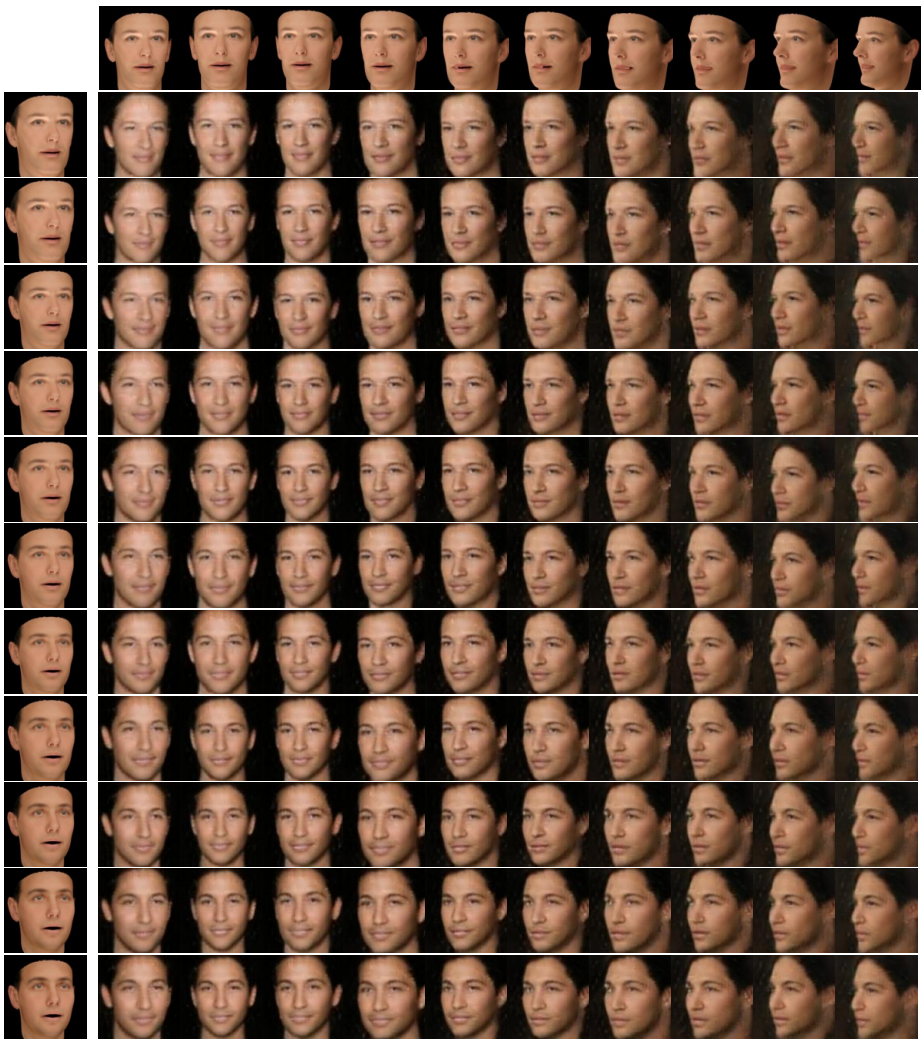

\hspace{1.2cm}
\foreach \x in {9,11,12,13,14,15,16,17,18,19}{\includegraphics[width=0.09\textwidth]{figures/supp/eccv_sup-108/00001/20_\x.jpg}}\\
\foreach \i in {20,18,16,14,12,10,08,06,04,02,01}{\includegraphics[width=0.09\textwidth]{figures/supp/eccv_sup-108/00001/\i_8.jpg}
\foreach \x in {9,11,12,13,14,15,16,17,18,19}{
\includegraphics[width=0.09\textwidth]{figures/supp/gen-eccv_sup-108/00001/\i_\x.jpg}\hspace{-0.08cm}}\\
}
\caption{Identity interpolation between first and second identities of the GANFaces dataset (Fig. 4 first two rows). Interpolation is done in the 3DMM space and projected onto realistic space by our framework. The vertical axis shows the identity interpolation under neutral lighting and expression with pose variation at the horizontal axis. Top-most and left-most 3DMM images indicate the respective identity and pose. Images in this figure are not included in the training.}
\label{fig:interpolation}
\end{figure*}

\begin{figure}
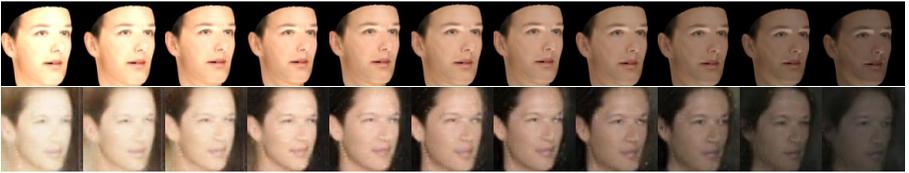

\center
\foreach \i in {16,15,14,13,12,11,10,09,08,07,06}{\includegraphics[width=0.09\textwidth]{figures/supp/eccv_sup-108/00003/\i_1.jpg}}\\
\foreach \x in {1}{
\foreach \i in {16,15,14,13,12,11,10,09,08,07,06}{
\includegraphics[width=0.09\textwidth]{figures/supp/gen-eccv_sup-108/00003/\i_\x.jpg}\hspace{-0.08cm}}\\
}
\caption{Effect of illumination changes to the generated images. Top row contains 3DMM synthetic images and the bottom contains the generated images by the framework given input images as the top row. Extreme lighting conditions result in blurry images as the real training set does not contain images of similar conditions.}
\label{fig:illumination}
\end{figure}

In this section, we show the generalization ability of our framework for unseen synthetic identities by interpolating in the identity space of our 3DMM model. Fig . \ref{fig:interpolation} shows how well shapes introduced by the 3DMM is learned so that the transition is smooth and accurate in the photorealistic space. The smooth transition between the two identities with pose variation also shows that the network did not overfit to the given synthetic data and is able to generate more photo-realistic images even without further training. Figure \ref{fig:illumination} shows that the framework also learned changes in the illumination strength and able to generate images with a controlled lighting variation.

\end{document}